%% file: Broyles.Hayner.ea.IROS2022a.tex
\documentclass[letterpaper, conference]{ieeeconf}

\IEEEoverridecommandlockouts
\usepackage{multicol}
\usepackage{amsmath,amssymb,amsfonts}
\usepackage{algorithmic}
\usepackage{graphicx}
\usepackage{siunitx}
\usepackage{textcomp}
\usepackage{xcolor}
\usepackage{hyperref}
\usepackage{pifont}
\usepackage{subcaption} 
\newcommand{\cmark}{\ding{51}}%
\newcommand{\xmark}{\ding{55}}%

\usepackage{array}
\newcolumntype{x}[1]{>{\centering\arraybackslash\hspace{0pt}}p{#1}}

\def\BibTeX{{\rm B\kern-.05em{\sc i\kern-.025em b}\kern-.08em
    T\kern-.1667em\lower.7ex\hbox{E}\kern-.125emX}}
    
\begin{document}
\input{arxiv_cover_page}

\title{WiSARD: A Labeled Visual and Thermal Image Dataset\\for Wilderness Search and Rescue}

\author{Daniel Broyles$^{\star,\dagger,1}$, Christopher R. Hayner$^{\star,1}$, Karen Leung$^{1,2}$
\thanks{$^\star$Both authors contributed equally.}
\thanks{$\dagger$ The views expressed in this
article are those of the author and do not reflect the official policy or position of the United States
Air Force, Department of Defense, or the U.S. Government.}
\thanks{$^1$Dept. of Aeronautics and Astronautics, University of Washington, USA. Emails: {\tt\small \{broyles, haynec, kymleung\}@uw.edu}, $^2$NVIDIA}%
\thanks{The Joint Center for Aerospace Technology Innovation (JCATI) provided funds to support this work. The authors would like to acknowledge Aadhar Chauhan, Matthew Klenk, Eli Pollack, Kenneth Wilsey, Will Browne, Evelyn Madewell, Timothy Zhou, Neil Gupta, Helen Kuni, and Karina Bridgman for their help in data collection and annotation. The authors would also like to acknowledge Dr. Juris Vagners and Linda Christianson.}
}

\newcommand{\klmargin}[2]{{\color{blue}#1}\marginpar{\color{blue}\raggedright\footnotesize [KL]: #2}}
\newcommand{\klnote}[1]{\textcolor{blue} {KL: #1}}
\newcommand{\chmargin}[2]{{\color{red}#1}\marginpar{\color{red}\raggedright\footnotesize [CH]: #2}}
\newcommand{\chnote}[1]{\textcolor{red} {CH: #1}}
\newcommand{\dbmargin}[2]{{\color{green}#1}\marginpar{\color{green}\raggedright\footnotesize [DB]: #2}}
\newcommand{\dbnote}[1]{\textcolor{green} {DB: #1}}
\newcommand{\postreview}[1]{\textcolor{black} {#1}}

\maketitle
\begin{abstract} 
Sensor-equipped unoccupied aerial vehicles (UAVs) have the potential to help reduce search times and alleviate safety risks for first responders carrying out Wilderness Search and Rescue (WiSAR) operations, the process of finding and rescuing person(s) lost in wilderness areas. 
Unfortunately, visual sensors alone do not address the need for robustness across all the possible terrains, weather, and lighting conditions that WiSAR operations can be conducted in. 
The use of multi-modal sensors, specifically visual-thermal cameras, is critical in enabling WiSAR UAVs to perform in diverse operating conditions. 
However, due to the unique challenges posed by the wilderness context, existing dataset benchmarks are inadequate for developing vision-based algorithms for autonomous WiSAR UAVs.
To this end, we present WiSARD, a dataset with roughly 56,000 labeled visual and thermal images collected from UAV flights in various terrains, seasons, weather, and lighting conditions.
To the best of our knowledge, WiSARD is the first large-scale dataset collected with multi-modal sensors for autonomous WiSAR operations. 
We envision that our dataset will provide researchers with a diverse and challenging benchmark that can test the robustness of their algorithms when applied to real-world (life-saving) applications.
Link to dataset: \url{https://sites.google.com/uw.edu/wisard/}
\end{abstract}

\section{Introduction}
Wilderness Search and Rescue (WiSAR) is the process of finding, assisting, and evacuating person(s) lost in wilderness areas. Search time is one of the most significant indicators of whether a Search and Rescue (SAR) operation results in rescue or a recovery. Oftentimes WiSAR operations are carried out in rugged and treacherous environments that may be difficult to search quickly or are unsafe for first responders. Naturally, there is great interest in using autonomous systems to help reduce search time and safety risks by providing imagery of the search area, including dangerous or hard-to-reach regions.

\begin{figure}[t]
    \centering
    \includegraphics[width=0.5\textwidth]{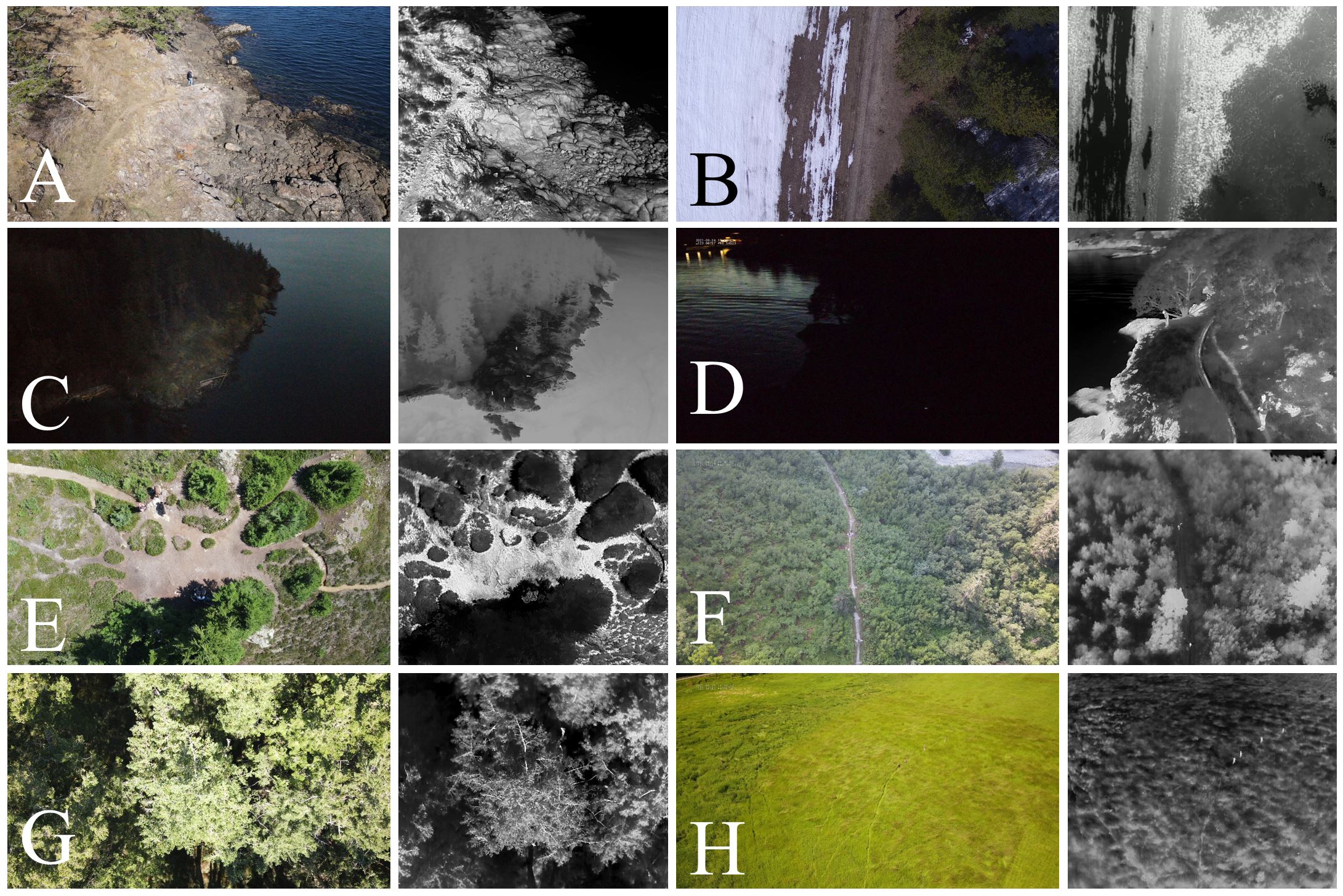}
    \caption{Examples of visual-thermal image pairs from WiSARD. The dataset encompasses a large diversity of wilderness settings: (A) Coastal, (B) Snow-Transition, (C) Dawn, (D) Dusk, (E) Hot Summer, (F) High-Altitude Viewpoint, (G) Forest, and (H) Open Field.}
    \label{fig:image diversity}
\end{figure}

Camera-equipped unoccupied aerial vehicles (UAVs) are used to augment aerial sensing roles that are typically carried out by human-occupied helicopters or small fixed-wing aircraft. Naturally, UAVs offer a cheaper, lighter, and safer alternative. However, UAVs face their own set of algorithmic challenges in wilderness settings. 
In order to detect humans in wilderness environments, the UAV must be able to \textbf{[C1]} deal with unstructured and rugged terrain, \textbf{[C2]} reason about occlusion due to foliage and shadows, \textbf{[C3]} understand the relative size of human objects from their high-altitude viewpoint, and \textbf{[C4]} be robust against diverse environments and weather conditions (e.g., snow, fog, forests, mountains, dawn, dusk). Visual and thermal images of a variety of wilderness environments are depicted in Figure~\ref{fig:image diversity}.

Unfortunately, \textit{visual} sensors alone do not address the need for robustness across all the possible terrains, weather, and lighting conditions that WiSAR operations can be conducted in. For instance, visual cameras are severely limited in low-light conditions (e.g., at dawn or dusk, see second row in Figure~\ref{fig:image diversity}). Noting that the autonomous vehicle (AV) community faces a similar set of challenges, the use of multi-modal sensors has proven to be necessary in providing autonomous systems with the capability to safely and reliably operate in such diverse settings. The different sensor modalities provide rich information when other modalities fail. For example, thermal cameras can still ``see'' in the dark.

While multi-modal sensor fusion and object detection are a very active research areas---largely driven by the AV community---many existing datasets and algorithms (e.g., \cite{GeigerLenzEtAl2013,waymo2019,nuscenes2019,TreibleSaponaroEtAl2017,ChoiKimEtAl2018}) do not adequately address the unique set of challenges characterizing autonomous WiSAR operations (see challenges C1--C4). In particular, the types of objects that an AV encounters are very different from what a WiSAR UAV would observe. Moreover, many existing AV datasets use a sensor suite that is impractical for WiSAR UAVs, and the viewpoint from a ground vehicle is completely different from a UAV's.
Due to reductions in size, weight, and cost, thermal cameras provide a viable solution to complement visual cameras on-board WiSAR UAVs.
Evidently, there is a clear gap in the existing field for a multi-modal dataset, specifically visual-thermal images, describing unstructured wilderness terrains in diverse operating conditions. Not only would a dataset be beneficial for the WiSAR community, but it would also serve as a unique dataset benchmark to develop robust algorithms which can operate in highly unstructured and diverse settings.

To this end, we are excited to present WiSARD (pronounced ``\textit{wizard}''), a labeled visual and thermal image dataset collected from UAV flights in various wilderness environments and operating conditions. The dataset includes three main subsets of data, the ``Visual Only'' set of 33,786 labeled visual images, the ``Thermal Only'' set of 22,156 labeled thermal images, and the ``Multi-modal'' set, a subset of the Visual and Thermal Only sets consisting of 15,453 temporally synchronized (i.e., taken at the same instant in time) visual-thermal image pairs.
To the best of our knowledge, WiSARD is the first large-scale dataset collected with multi-modal sensors for autonomous WiSAR operations. We envision that WiSARD will be a useful benchmark to further push computer vision and robotics research, especially with potentially high impact for WiSAR applications.
WiSARD contains a diverse set of images, covering different terrains, lighting conditions, and weather/seasons (see Figure~\ref{fig:image diversity} for examples), and therefore presents interesting challenges for many computer vision tasks. For instance, the visual and thermal images can look vastly different in poor lighting conditions or in the snow, thus making feature matching and thermal-visual image registration extremely challenging. Moreover, the diversity aspect of WiSARD could (i) make it a useful benchmark for testing the robustness of computer vision models, (ii) present an interesting transfer learning benchmark, and (iii) enable deeper understanding of distributional shifts in machine learning.

\noindent\textbf{Organization: } In Section~\ref{sec:uniqueness of dataset}, we discuss the uniqueness of WiSARD by highlighting key desiderata and limitations in existing datasets. Then we describe our data collection process, including a breakdown of the diverse makeup of the dataset, in Section~\ref{sec:data collection}, and our labeling procedure in Section~\ref{sec:labeling}. In Section~\ref{sec:interesting samples}, we showcase some baseline object detection results as well as a few representative challenging examples and discuss why many existing algorithms would perform poorly on our dataset. Finally, we conclude and lay out future research directions in Section~\ref{sec:conclusion and future direction}.

\begin{table*}[t]
    \vspace{3mm}
    \caption{Comparison of publicly available datasets related to WiSARD. }
    \label{tab:datasets}
    \centering
    \begin{tabular}{|x{4cm}|c|c|c|c|c|c|}
        \hline
        \textbf{Dataset} & \textbf{Domain} & \textbf{Task} & \textbf{Environment} & \textbf{Visual} & \textbf{Thermal} & \textbf{Viewpoint} \\ \hline\hline
        INTERACTION \cite{ZhanSunEtAl2019} & AD & Trajectory forecasting & Roads & \cmark & \xmark & Aerial\\ \hline
        KAIST multi-spectral$^*$ \cite{ChoiKimEtAl2018}, \cite{Hwang2015peddet} & AD & Object detection \& more & Roads & \cmark & \cmark &  Ego-centric\\ \hline
        KITTI$^*$ \cite{GeigerLenzEtAl2013}, Waymo Open Dataset$^*$ \cite{waymo2019}, nuScenes$^*$ \cite{nuscenes2019} & AD & Object detection \& more & Roads & \cmark & \xmark &  Ego-centric \\ \hline
        CATS \cite{TreibleSaponaroEtAl2017}& Various & Stereo matching & Various & \cmark & \cmark &  Various \\ \hline
        VisDrone \cite{ZhuWenEtAl2021} & Surveillance & Object detection \& tracking & Urban \& rural cities & \cmark & \xmark &  Ego-centric \\\hline
        CAMEL \cite{GebhardtWolf2018} & Surveillance & Object detection \& tracking & Urban & \cmark & \cmark & Stationary tripod\\ \hline
        MultiPoint \cite{AchermannKolobovEtAl2020} & Mapping & Visual-thermal registration & Farmland & \cmark & \cmark & Ego-centric \\ \hline
        HERIDAL \cite{Bozic-StulicMarusicEtAl2019} & SAR & Object detection & Wilderness & \cmark & \xmark &  Ego-centric \\ \hline
        UMA-SAR$^*$ \cite{MoralesVazquezEtAl2021} & SAR & Unspecified & Disaster sites & \cmark & \cmark & Stationary tripod\\
        \hline\hline
        \textbf{WiSARD (Ours)} & \textbf{SAR} & \textbf{Object Detection} & \textbf{Wilderness} & \cmark & \cmark & \textbf{Ego-centric} \\ \hline
    \end{tabular}
    \vspace{-2mm}
    \begin{flushleft}
    {\footnotesize{$\: ^*$Also contains LiDAR, Stereo RGB, and/or GPS/IMU sensor readings too.}}\\
    {\footnotesize{$\:$ AD is short for autonomous driving.}}
    \end{flushleft}
    \vspace{-8mm}
\end{table*}

\section{Uniqueness of WiSARD}
\label{sec:uniqueness of dataset}
We first highlight key desiderata that we believe are critical for enabling and accelerating the development of computer vision algorithms for WiSAR UAVs. Then we provide an overview of existing datasets, and show that WiSARD uniquely addresses all the desiderata.

\subsection{Key desiderata} 

We worked with first responders to understand important practical considerations for WiSAR UAVs, and identified the following desiderata as necessary in ensuring that a dataset will be useful for developing vision-based autonomy for WiSAR missions.

\noindent\textbf{[D1] Operational representative environment}: Images need to be of \textit{wilderness} environments representative of where WiSAR operations may take place (e.g., in forests or mountainous areas). Wilderness environments are distinctly different from urban settings.

\noindent\textbf{[D2] Multi-modal sensors}: Reliance on only visible spectrum imaging sensors is not sufficient, especially in low light situations (i.e. night, dawn, dusk). The introduction of a thermal cameras provides a complementary sensing modality when other modalities suffer.

\noindent\textbf{[D3] Weather and environment diversity} WiSAR operations can occur anywhere and at any time. As such, the dataset imagery should encompass a large variety of weather conditions and environments (e.g., different seasons, locations, time of day). Even for the same location, the terrain can look vastly different depending on the season.

\noindent\textbf{[D4] Ego-centric viewpoint}: The data viewpoint should match the system's viewpoint during deployment such that any algorithms developed using the dataset can be directly deployable on a UAV. An anti-example: trajectory data of cars traveling on a highway collected from aerial imagery \cite{ZhanSunEtAl2019} may be useful for driving behavior prediction research, but the same information/data representation may not be necessarily realizable during real-world deployment.

\noindent\textbf{[D5] Practical sensor suite}:  We envision that technologies stemming from WiSAR research should be beneficial and accessible for all. Therefore the dataset should be collected from practical and commercial off-the-shelf sensors. We hope this requirement reduces friction for researchers and practitioners from transferring their algorithms into practice.

\subsection{Related datasets}
In recent years, there has been tremendous progress made in computer vision research, accelerated by the abundance of large-scale freely accessible datasets. Consequently, there have been numerous efforts from academia and industry across the globe in gathering and curating datasets to share with the community. In this section, we give a brief overview of related datasets and describe why algorithms developed on those datasets cannot easily transfer to the WiSAR setting. Table~\ref{tab:datasets} provide a summary of related datasets and the columns refer to some of our key desiderata. 

\noindent\textbf{Datasets for object detection:} Many large-scale datasets for object detection tasks exist (e.g., COCO \cite{LinMaireEtAl2015}, ImageNet \cite{DengDongEtAl2009}, and Open Image Dataset \cite{OpenDataset2020}), but unfortunately are not suitable for developing human detection models in wilderness environments. Many of these types of datasets contain images of a large variety of objects (e.g., household items) which are often clearly visible in the image. Moreover, the images were collected from photo galleries on the internet and thus have inconsistent viewpoints. While in the WiSAR settings, the opposite occurs---images are taken from a high-resolution camera mounted on a UAV flying at high altitudes. From this viewpoint, any humans in-frame would make up only a very small portion of the image and could be partially occluded by foliage or shadows.
In terms of datasets containing imagery from an aerial perspective, the VisDrone dataset \cite{ZhuWenEtAl2021} is large-scale benchmark dataset that is similar to WiSARD in the inclusion of imagery of human subjects from an aerial perspective, but it is limited to urban and rural environments. In general, these environments are more structured (e.g., straight lines on roads, well-defined features) which do not easily transfer to less structured wilderness environment (e.g., lack of straight lines, diffused features).
Essentially, the rugged wilderness terrain presents very different textures and features typically found in urban or rural environments.
In short, due to the unique characteristics of wilderness environments, algorithms developed using large-scale generic object detection datasets or different view points would not transfer well to detecting human(s) in rugged terrains.

\noindent\textbf{Datasets for multi-modal fusion:} 
While there has been huge progress made in multi-modal sensor fusion algorithms---largely propelled by autonomous driving research---the wilderness setting presents unique challenges that make existing multi-modal sensor datasets inadequate for WiSAR applications. Popular multi-modal sensor dataset stemming from autonomous driving applications (e.g., KITTI \cite{GeigerLenzEtAl2013,GeigerLenzEtAl2012},  Waymo Open Dataset \cite{waymo2019} and nuScenes \cite{nuscenes2019}) often do not include thermal images and instead include LiDAR readings, thus sensor fusions algorithms developed on those datasets cannot apply to our WiSAR setting. There are other multi-modal datasets that include thermal images such as \cite{TreibleSaponaroEtAl2017,ChoiKimEtAl2018} for autonomous driving applications and \cite{GebhardtWolf2018} for pedestrian tracking. We hope that sensor fusion algorithms developed on those datasets can extend to the WiSAR setting, but we note that the varied lighting conditions and rugged terrain in WiSAR operations make fusing visual and thermal images extremely challenging. Regardless, many existing datasets are taken from ground vehicles which makes the transfer to an aerial view challenging.
Algorithms developed on the visual-thermal dataset presented in \cite{AchermannKolobovEtAl2020} which was taken from a UAV flying over Swiss farmland in broad daylight could be useful for WiSAR settings, but would not be very robust to varied weather and lighting conditions that are characteristic of WiSAR operations.

\noindent\textbf{Datasets for search and rescue:} Unfortunately, there are not many existing datasets targeted for search and rescue applications. To the best of our knowledge, the HERIDAL dataset \cite{Bozic-StulicMarusicEtAl2019} is the most similar in spirit to WiSARD---both are datasets of images taken from a UAV flying over wilderness environments. The main difference is that WiSARD consists of visual and thermal images whereas the HERIDAL dataset only contains visual images. Additionally, the scale of the HERIDAL dataset is much smaller than WiSARD, with approximately 500 labeled visual images compared to the over 10,500 visual-thermal image pairs in WiSARD. Consequently, WiSARD contains a larger diversity of lighting and weather conditions.
The UMA-SAR dataset \cite{MoralesVazquezEtAl2021} is another dataset targeted for search and rescue operations. The dataset contains raw sensor readings from visual and thermal cameras, LiDAR, IMU, and GPS mounted on a ground vehicle following first responders during an outdoor search and rescue exercise. Although the application is relevant, the environment and viewpoint make this data unusable for our intended UAV-aided WiSAR operations. Only the raw sensor readings are provided, thus the lack of ground truth annotations limits the utility of the dataset.

\begin{figure}[h!]
    \centering
    \includegraphics[width=0.5\textwidth]{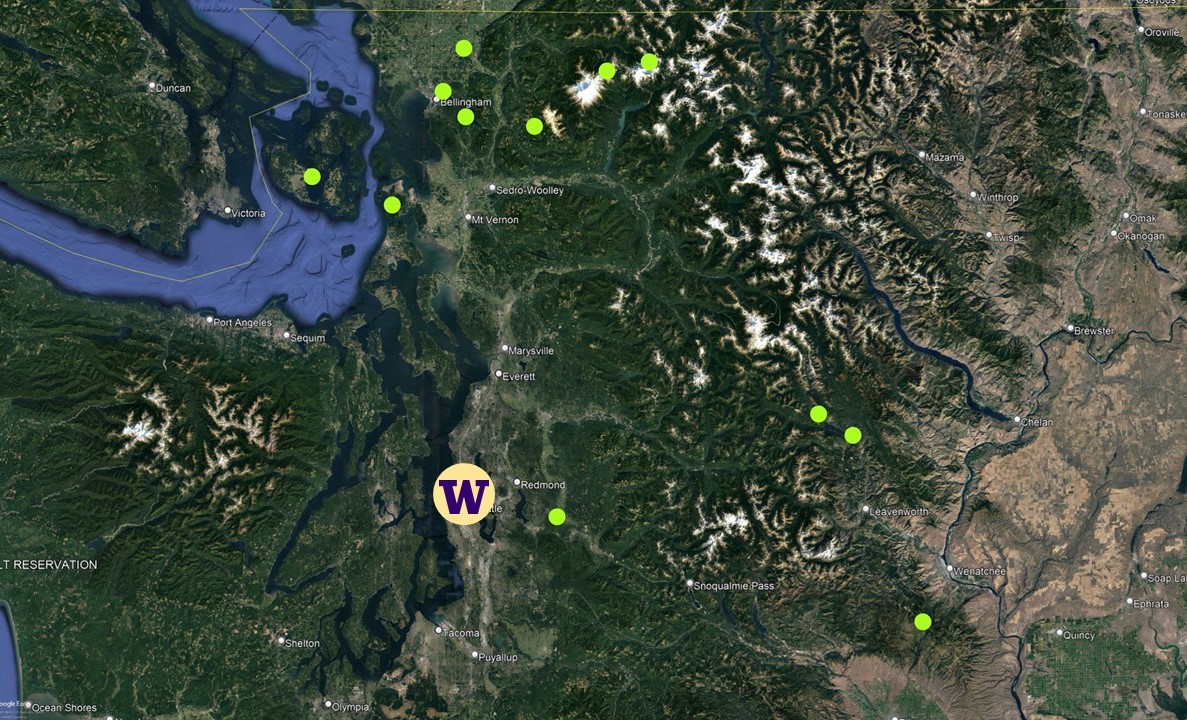}
    \caption{Green dots represent locations in Washington State, USA, where WiSARD imagery was collected \cite{GoogleEarthImage}.\\}
    \label{fig:washmap}
\end{figure}

\section{Data Collection Process}
\label{sec:data collection}
In this section, we describe the hardware used to collect the images, and the data collection process, including details regarding location, terrain, weather, and time diversity.
In total, the dataset is comprised of 55,942 images taken from a UAV flying in 12 different locations, marked in Figure~\ref{fig:washmap}, and at different times of the day/night and year.

\subsection{Hardware and sensor suite}
\begin{figure}
    \centering
    \includegraphics[width=0.5\textwidth]{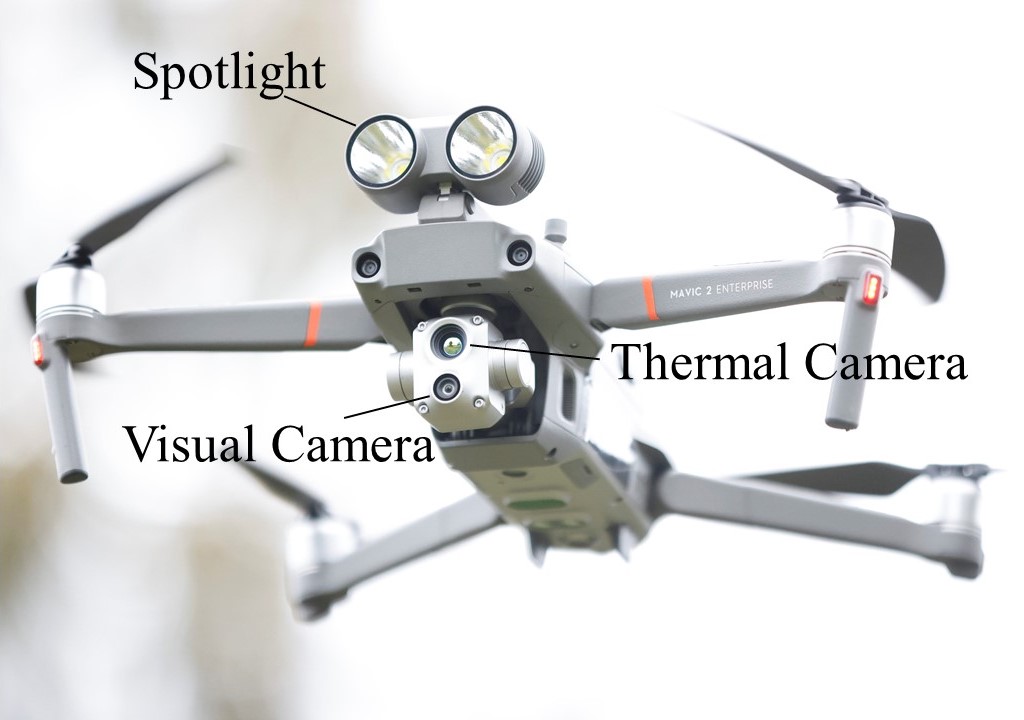}
    \caption{A DJI Mavic 2 Enterprise Advanced with Spotlight was used to collect visual and thermal images in a variety of operating conditions.\\}
    \label{fig:hardware}
\end{figure} 
We used a variety of camera-equipped UAVs to collect video imagery in different wilderness environments, thereby addressing desideratum [D4]. All images were collected using commercially available UAVs and sensors, therefore addressing desideratum [D5]. Table \ref{hardware} lists all the UAV and sensor configurations used during collection of WiSARD. The DJI Mavic 2 Enterprise Advanced, one of the primary UAVs used for this effort, is shown in Figure~\ref{fig:hardware}.

\begin{table}[t]
\vspace{3mm}
\caption{UAV Hardware Configurations}
\centering
\begin{tabular}{|c|c|c|c|}
\hline 
\textbf{UAV} & \textbf{Sensor}& \textbf{Modality(s)}& \textbf{Resolution$^{\mathrm{a}}$} \\
\hline \hline
DJI Mavic 2  & Internal & Visual & 3840$\times$2160$^{\mathrm{b}}$  \\
Enterprise Advanced &  & Thermal$^\mathrm{c}$ & 640$\times$512  \\
\hline
DJI Matrice  & FLIR Duo & Visual & 3840$\times$2160  \\
600 Pro & Pro R  & Thermal$^\mathrm{c}$ & 640$\times$512  \\
\hline
DJI Inspire 1 v2 & Zenmuse Z3 & Visual & 3840$\times$2160  \\
\hline
DJI Mavic Mini & Internal & Visual & 2720$\times$1530  \\
\hline
DJI Phantom 4 Pro & Internal & Visual & 3840$\times$2160  \\
\hline
\end{tabular}
\vspace{-1mm}
\begin{flushleft}
{\footnotesize{$^{\mathrm{a}}$Resolution given in pixels.  $\: ^{\mathrm{b}}$Some imagery in 1920$\times$1080.  $\: ^{\mathrm{c}}$Long Wave Infrared}}\\
\end{flushleft}
\label{hardware}
\end{table}

\subsection{\postreview{Data collection flights}}
\postreview{ At each location, camera-equipped UAVs were deployed to collect videos of volunteer human subjects scattered throughout the environment with altitudes ranging from $\SI{20}{\meter}$ to $\SI{120}{\meter}$. Flight profiles for video collection included a mix of moving ($<\SI{10}{\meter/\second}$) and stationary (hover), and the aspect angle of the camera with respect to the human subjects was varied. As a result, some motion blur is present in the dataset. The thermal cameras performed Flat Field Corrections throughout each flight \footnote{A Flat Field Correction corrects for variations in pixel sensitivity and results in a few dropped frames \cite{seibert1998flat}
}. All video was recorded at \SI{30}{\hertz} \footnote{The manufacturer claims that the visual imagery from the DJI Mavic 2 Enterprise Advanced was at \SI{30}{\hertz}, however in our testing we found this closer to \SI{29.95}{\hertz} resulting in a one or two image difference between corresponding multi-modal image sets}, and every sixth frame was retained for annotation. We additionally provide all of the raw unannotated video.}

\subsection{Imagery statistics}
The guiding principal in the collection of WiSARD was to remain relevant to the real-world WiSAR scenario. As WiSAR missions are often carried out in a wide variety of terrains, seasons, and times of day (dawn, day, dusk, night), it is important that we too curated images that captured the multitude of settings.
To ensure that we address desideratum [D1], we consulted first responders from the State of Washington, USA, to identify data collection sites that are representative of operational environments.

\subsubsection{Terrain diversity}
We are fortunate that the State of Washington is unique in its inclusion of a high variability of environments and vegetation. While we predominantly collected data from the Western Washington region (west of the Cascade Mountains), we were still able to capture a diverse set of terrain features, such as forests, fields, rocky regions, coastal regions, and areas with full or partial snow coverage. Examples of terrain diversity are shown in Figure~\ref{fig:image diversity}, and the distribution of the different terrain features in WiSARD is presented in Table~\ref{tab:terrain diversity}. 

\begin{table}[h!]
\centering
\vspace{3mm}
\caption{Wilderness Terrain Distribution}
\begin{tabular}{|c|c|c|c|c|c|}
\hline
&\textbf{Forest} & \textbf{Fields}& \textbf{Rocky}& \textbf{Coastal} &\textbf{Snow} \\
\hline\hline
Total images & 36,384 & 16,431 & 11,555 & 12,353 & 9,011 \\
\hline
Visual images & 16,360 & 3,996 & 4,547 & 5,212 & 2,137 \\
\hline
Thermal images & 20,024 & 12,435 & 7,008 & 7,141 & 6,874 \\
\hline 
Multi-modal pairs & 14,711 & 0 & 5,361 & 7,022 & 0 \\
\hline
\end{tabular}
\label{tab:terrain diversity}
\end{table}

In addition to terrain diversity, human subject locations within the environment were varied. Human subjects were imaged on trails, off trails, and on dirt roads. The distribution of human subject locations is presented in Table~\ref{tab:human placement}.

\begin{table}[h]
\centering
\vspace{3mm}
\caption{Human Location Distribution}
\begin{tabular}{|c|c|c|c|}
\hline
&\textbf{On trail} & \textbf{Off trail}& \textbf{Dirt road} \\
\hline\hline
Total images & 14,888 & 41,217 & 6,850\\
\hline
Visual images & 7,701 & 12,507 & 4,857 \\
\hline
Thermal images & 7,187 & 28,710 & 1,993 \\
\hline 
Multi-modal pairs & 9,401 & 16,257 & 0 \\
\hline
\end{tabular}
\label{tab:human placement}
\end{table}

\subsubsection{Seasonal diversity}
We also collected images during different times of the year to capture seasonal diversity; each season poses a different set of unique computer vision challenges. For example, in snowy conditions, white pixels will be more prominent resulting in a potentially feature-poor visual image, however the contrast of a human's thermal signature against the cold snow will result in more pronounced features. Examples of seasonal diversity are shown in Figure~\ref{fig:image diversity}, and the distribution of the different seasons in WiSARD is shown in Table~\ref{tab:seasonal diversity}.
\begin{table}[h]
    \centering
    \vspace{3mm}
    \caption{Seasonal Diversity Distribution}
    \begin{tabular}{|c|c|c|c|c|}
    \hline
        & \textbf{Summer} & \textbf{Autumn} & \textbf{Winter} & \textbf{Spring}  \\
        \hline\hline
        Total images & 6,310 & 11,218 & 13,303 & 23,654\\
        \hline
        Visual images & 1,468 & 3,120 & 3,282 & 14,286 \\
        \hline
        Thermal images & 4,842 & 8,098 & 10,021 & 9,368 \\
        \hline
         Multi-Modal pairs & 0 & 4,192 & 2,293 & 13,059 \\
        \hline
    \end{tabular}
    \label{tab:seasonal diversity}
\end{table}

\subsubsection{Lighting diversity}
\label{subsubsec:lighting diversity}
Since WiSAR missions can occur anytime in the day or night, we also collected images in different lighting conditions. Different lighting conditions pose different sets of computer vision challenges. For example, at night, visual cameras are essentially useless without the aid of a light source, whereas the detection task using thermal images is particularly challenging during dusk and dawn due to the effect known as ``thermal crossover''   \cite{Gurton_undated-sg}. Thermal crossover is the phenomenon specific to infrared spectrum sensors where the thermal signatures of two distinct objects are indistinguishable (e.g., a car blending into the road). This loss of contrast is the result of temperature conditions, emissivity of the objects, and sensitivity of the sensor, and can occur at any time, but most commonly occurs during dusk and dawn \cite{Zhao2016-gy}. Table~\ref{tab:lighting diversity} lists the distribution of lighting conditions found in WiSARD.
\begin{table}[h]
    \centering
    \vspace{3mm}
    \caption{Lighting Diversity Distribution} 
    \begin{tabular}{|c|c|c|c|c|}
    \hline
         & \textbf{Day} & \textbf{Night} & \textbf{Dawn} & \textbf{Dusk}  \\
        \hline\hline
        Total images & 35,695 & 7,141 & 1,753 & 9,896 \\
        \hline
        Visual images & 16,207 & 1,145 & 1,378 & 3,426\\
        \hline
        Thermal images & 19,488 & 5,996 & 375 & 6,470\\
        \hline
        Multi-Modal pairs & 10,390 & 2,293 & 747 & 6,114\\
        \hline
    \end{tabular}
    \label{tab:lighting diversity}
\end{table}

Even in broad daylight, the poor lighting condition below the tree canopy presents a challenge. The amount of light passing through the canopy depends on the density of the tree cover which in turn can vary seasonally. As such, we also captured imagery from below the tree canopy as well as above when in a forest setting which is shown in Table~\ref{tab:camera perspective}. 

\begin{table}[h]
\centering
\vspace{3mm}
\caption{Camera View Diversity Distribution}
\begin{tabular}{|c|c|c|c|c|}
\hline
&\textbf{Above canopy} & \textbf{Below canopy} & \textbf{No canopy}\\
\hline\hline
Total images & 35,253 & 2,924 & 16,308 \\
\hline
Visual images & 14,777 & 2,480 & 4,899 \\
\hline
Thermal images & 20,476 & 444 & 11,409 \\
\hline 
Multi-modal pairs & 15,616 & 888 & 3,040 \\
\hline
\end{tabular}
\label{tab:camera perspective}
\end{table}

\subsubsection{Human Subject diversity}
It was critical that we vary the human subjects to remove as much bias from our dataset as possible. To this effect, we directed human subjects to change their clothing and accessories throughout a data collection flight (e.g., shirt color, long/short sleeve, head gear). The human subjects also assumed different body stationary positions such as sitting, laying down, and standing as well as dynamic actions such as walking, running, and waving at the UAV. In addition to human subject clothing, positions, and actions, we made efforts to vary the actual human subjects that make up WiSARD. The age range of volunteer human subjects in our dataset is from 11 to 70 years old. However, since many of the human subjects were sourced from undergraduate and graduate population, there is bias towards the 18-25 year old age range. While having diverse skin tones in human subjects is important for facial recognition tasks, since the size of our human subjects in the images are small, skin-tones do not offer any visible features when compared to clothing color, hair color, or head gear (e.g. hats, helmets).

\section{Data Labeling}
\label{sec:labeling}

In this section, we describe the labeling and annotation process of our dataset.

\subsection{Ground truth data}

For image annotation we use Supervisely, an online annotation platform that provides distributed annotation through a web interface as well as tracking algorithms which assist annotators by interpolating bounding boxes between frames. The annotator then manually adjusts the bounding box if necessary. 
The bounding box information is encapsulated in a \texttt{.txt} file in following format.
\begin{equation}
    \begin{bmatrix}
    \text{\textbf{C}} & \text{\textbf{X}} & \text{\textbf{Y}} & \text{\textbf{W}} & \text{\textbf{H}}
    \end{bmatrix}.
    \label{eq:annotation} 
\end{equation}
In Equation \ref{eq:annotation},
\textbf{C} pertains to the object class, since we only consider the human class, this entry is set to 0 for all annotations. \textbf{X} and \textbf{Y} are the $x$ and $y$ coordinates of the center of the bounding box relative to the upper left corner of the image. \textbf{W} and \textbf{H} are the bounding box width and height relative to the image width and height. Human shadows were intentionally not included in the bounding boxes as they are often not visible in the thermal modality. Note that we made efforts to include imagery which has non-human subject examples present such as a domestic dog and cat. These will aid in reducing the number of False Positive detections on wildlife which is likely to be encountered in a real WiSAR operation. Due to safety and logistical reasons, we were unable to deliberately image any natural wildlife.

\subsection{Metadata}
We also provide data about the flight conditions associated with each subset of images. We hope that this information may be useful when studying domain transfer of algorithms when applied to different settings. For instance, developing algorithms trained on forest images to perform well on coastal images.
Metadata tags are included in a \texttt{.yaml} file for each folder and the metadata applies to all of the images within the folder. We include the following tags:
\vspace{2mm}

\noindent \textbf{Terrain:} \textit{Forest, Fields, Rocky, Coastal,} \textit{Snow}

\noindent \textbf{Time-of-day:}  \textit{Day, Night, Dawn,} \textit{Dusk}

\noindent \textbf{Human Location:} \textit{On trail, Off trail,} \textit{Dirt Road}

\noindent \textbf{Season:} \textit{Summer, Autumn, Winter, Spring}

\noindent \textbf{Camera View:} \textit{Above Canopy, Below Canopy,} \textit{No Canopy}

\vspace{2mm}
Note that, with the exception of the terrain category, these tags are mutually exclusive such that only one designation per category is allowed for each image. 

\section{Baseline Results and Representative Challenging Examples}
\label{sec:interesting samples}
\postreview{In this section, we present baseline results from object detection models trained on WiSARD, and consider a few representative examples to showcase why the WiSARD is a challenging benchmark to test computer vision and sensor fusion algorithms on}. In Sections \ref{subsec:poor lighting}, \ref{subsec:partial occlusions} and \ref{subsec:rocky terrain}, we discuss cases that are particularly challenging for the object detection task, and in Section~\ref{subsec:registration challenge}, we discuss the challenges faced in a visual-thermal image registration task.

\begin{table}[t]
    \centering
    \vspace{3mm}
    \caption{Object detection results using baseline model trained on WiSARD. (Higher is better.)}
    \begin{tabular}{|x{1cm}|x{1.6cm}|x{1cm}|c|c|}
    \hline
        {\bf Dataset} & \textbf{mAP} (0.5:0.05:0.95) & \textbf{mAP} (0.5) & \textbf{Precision}& \textbf{Recall}  \\
        \hline\hline
        Visual & 0.546 & 0.892 & 0.944 & 0.845\\
        \hline
        Thermal & 0.572 & 0.977 & 0.976 & 0.959\\
        \hline
    \end{tabular}
    \label{tab:baseline results}
\end{table}

\subsection{\postreview{Baseline object detection results}}
\begin{figure}[t] 
\centering
   \includegraphics[width=\linewidth]{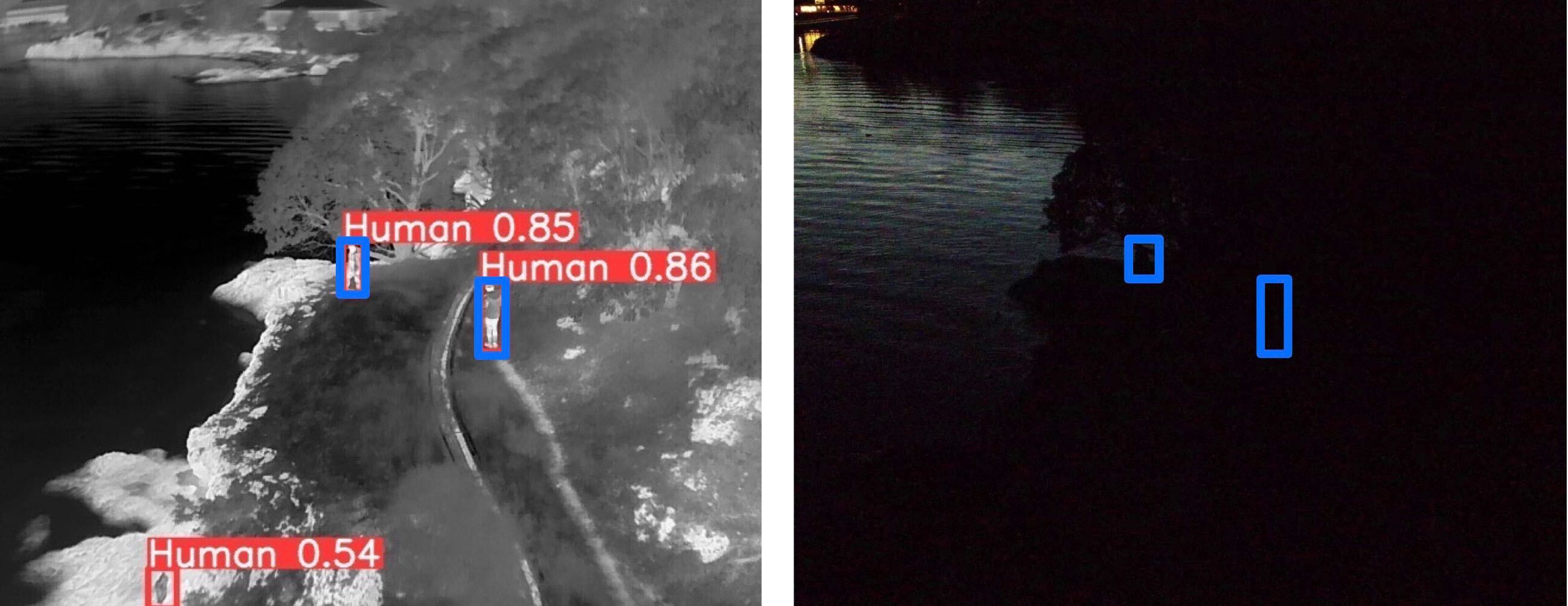}
   \caption{An example with low-light dusk conditions where the visual object detection model suffers in comparison to the thermal. Visual (right) and thermal (left) processed images; YOLOv5 detections in red, ground truth boxes in blue. Note that in the detection in the lower left of the thermal image is a False Positive.}
   \label{fig:low light comparison}
\end{figure}

We present baseline results generated from using a common real-time object detector trained on WiSARD, and this is presented in Table~\ref{tab:baseline results}. Specifically, we use YOLOv5 \cite{yolov5} as it is a well known and commonly used state-of-the-art object detector and serves as a good baseline\footnote{YOLOv5 is largely based on the YOLOv3 architecture \cite{yolov3}} \postreview{We trained both a Visual Only and Thermal Only model which were used to obtain the results shown in Table~\ref{tab:baseline results} and Figures \ref{fig:low light comparison}, \ref{fig:occluded comparison}, and \ref{fig:temp gradient comparison}. The models were trained on WiSARD using a 60-20-20 split for training, validation, and test respectively.} Additionally, we tiled the visual imagery to $512 \times 512$ pixels and culled the visual images prior to training using the methods described in \cite{HaynerGuptaEtAl2021}. \postreview{We tiled the visual imagery because our dataset includes several different visual resolutions and tiling them asserts a common image resolution across them. Additionally, we culled the tiles which did not have a human subject in them to ensure a well-balanced dataset during training and testing.\footnote{We kept in $0.1\%$ of tiles without a human subject to leave in true negative examples in the train, validation, and test dataset.} We did not tile the thermal imagery as all the thermal cameras used provided the same resolution imagery. Additionally the thermal resolution is significantly less than the visual resolution and the resultant tiles would be too small for the object detection task. Furthermore, \cite{OzgeUnelBurakEtAl2019} demonstrated that, for small object detection, tiling and culling improves detection performance.}

\postreview{In these baseline results, the Thermal Only model outperforms the Visual Only model across all metrics. One possible reason for this is due to a bias in the dataset towards forested imagery, in which the thermal signatures of humans below the tree canopy may exhibit stronger contrast compared to its visual counterpart. However, in general, environmental factors play a large role in the ability of either model to detect humans, and in some cases the Visual Only model outperforms the Thermal Only model by a large margin (see Figure \ref{fig:temp gradient comparison}).}

\postreview{In determining relevant metrics, we first define the situational analogues to False Positives and False Negatives which are \textit{detecting a non-human as a human} and \textit{not detecting a human}, respectively. False Negatives are far less acceptable than False Positives as it takes little effort on the part of the operator to visually verify the correctness of a detection whereas missing a detection results in delaying the rescue or worse. With this in mind, we use the common object detection performance metrics of Precision and Recall, defined as follows,}
\begin{equation}
\text{Precision} = \frac{\text{True Positives}}{\text{True Positives} + \text{ False Positives}},
\label{eqn:precision}
\end{equation}
\begin{equation}
\text{Recall} = \frac{\text{True Positives}}{\text{True Positives} + \text{False Negatives}}.
\label{eqn:recall}
\end{equation}

\postreview{In terms of performance evaluation, since False Negatives are more detrimental to the search and rescue operation, we emphasize that the Recall metric is more important than the Precision metric.}

\postreview{Additionally, we note that anything more then a rough estimation of the human subjects' position within frame will have diminishing returns as human operators could (and often will) monitor the detections in real-time and localize the human subject as required for the search and rescue team.}
We consider Mean Average Precision (mAP) using the convention introduced by the COCO dataset \cite{LinMaireEtAl2015}. We report both the mAP@0.5 and mAP@0.5:0.05:0.95 metrics, but since a rough estimate of the human subject's position is adequate, considering the mAP@0.5 metric is sufficient.

A complete list of our baseline metrics, as well as the associated code, \postreview{and imagery for reproducing these results} can be found on our accompanying WiSARD website.

\subsection{Poor lighting conditions} 
\label{subsec:poor lighting}
WiSAR teams operate around the clock and, in many cases, the search for a lost or missing individual must continue even after the sun has gone down or during hours when lighting conditions are not ideal for computer vision-related tasks. 
An example of a low-light dusk condition is shown in Figure~\ref{fig:low light comparison}. The detection performance of the YOLOv5 model trained on WiSARD is also shown; as expected, the Visual Only model does not detect the two human subjects in the scene, however the Thermal Only model detects two humans subjects and one False Positive on a human-sized branch.

\subsection{Partial occlusions} 
\label{subsec:partial occlusions}
\begin{figure}[t] 
\centering
   \includegraphics[width=\linewidth]{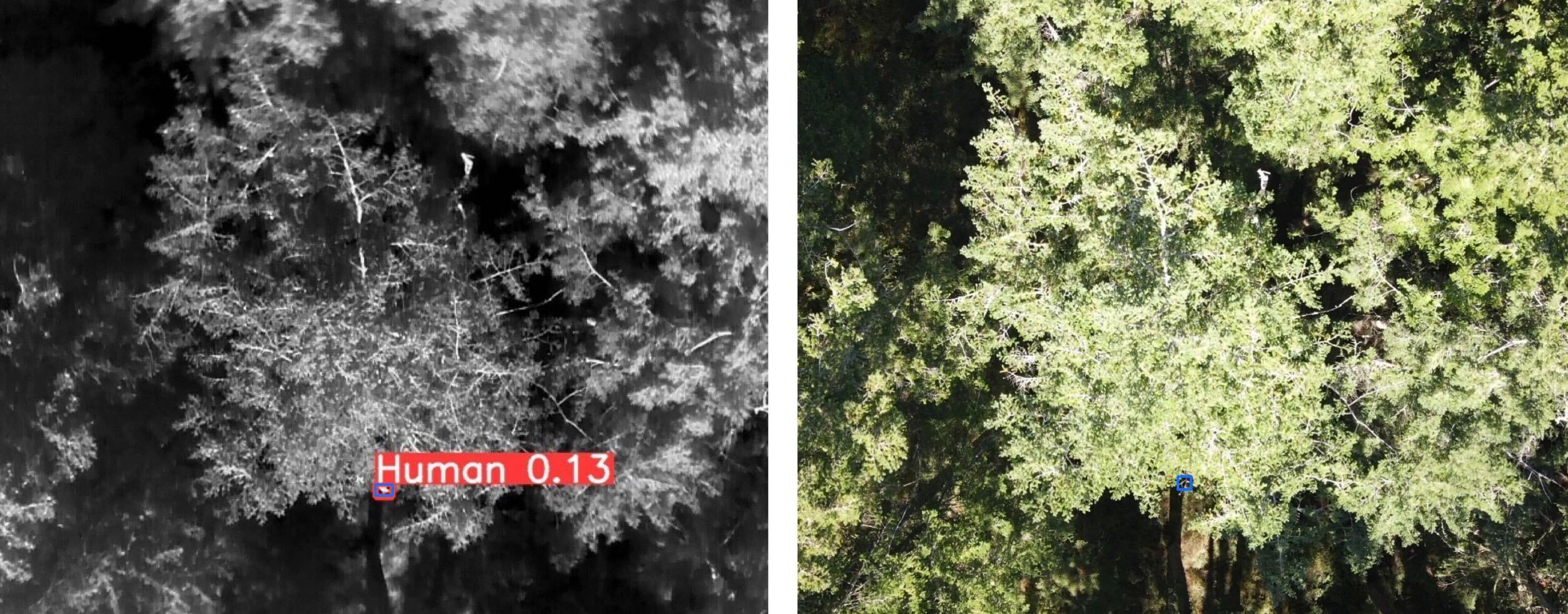}
   \caption{An example with a partially occluded human subject correctly detected by the Thermal Only model but missed by the visual. Visual (right) and thermal (left) processed images; YOLOv5 detections in red, ground truth boxes in blue.}
   \label{fig:occluded comparison}
\end{figure}

Occlusion from vegetation or terrain features while conducting an aerial search is common in WiSAR operations. From a computer vision perspective, the detection task is significantly harder due to missing features normally present in unoccluded images. In Figure \ref{fig:occluded comparison}, the Thermal Only model identifies the partially occluded human subject, but the Visual Only model misses the partially occluded subject completely.

\subsection{Thermal crossover}
\label{subsec:rocky terrain}
\begin{figure}[t] 
\centering
   \includegraphics[width=\linewidth]{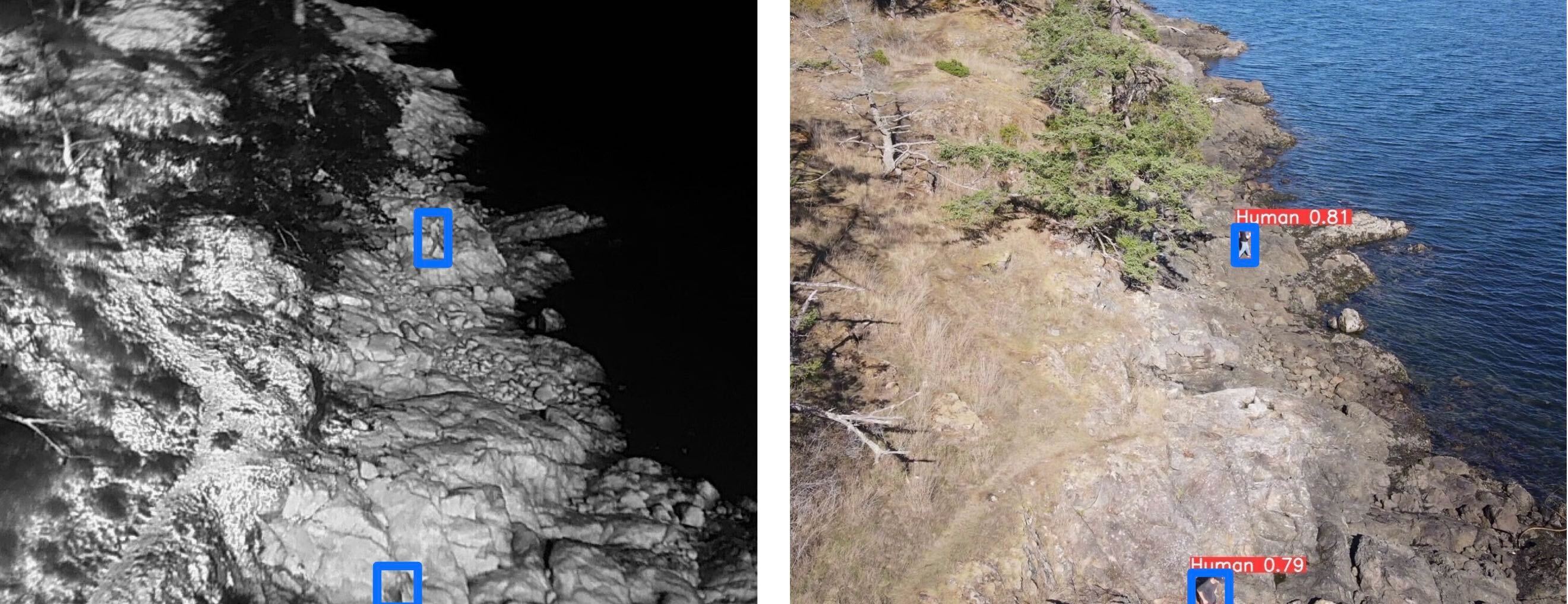}
   \caption{An example with rocky terrain which causes False Positive and False Negative detections in the Thermal Only model. Visual (right) and thermal (left) processed images; YOLOv5 detections in red, ground truth boxes in blue.}
   \label{fig:temp gradient comparison}
\end{figure}

The thermal modality is especially susceptible to ambient temperatures. In cases where the background is at a similar temperature to the human subject, there are very few distinguishing features to extract and make a detection. This commonly occurs when the human subjects are on rocks or sand which have a greater thermal mass then grass or dirt, or during periods of thermal crossover (as described in Section~\ref{subsubsec:lighting diversity}). A coastal scene featuring human subjects on rocky terrain is shown in Figure~\ref{fig:temp gradient comparison}. In this case, the Visual Only model correctly detects the human with high confidence. The Thermal Only model misses the humans completely because the the contrast between the humans' thermal signature and the surrounding environment is significantly low. 

\begin{figure}[t] 
\centering
   \includegraphics[width=\linewidth]{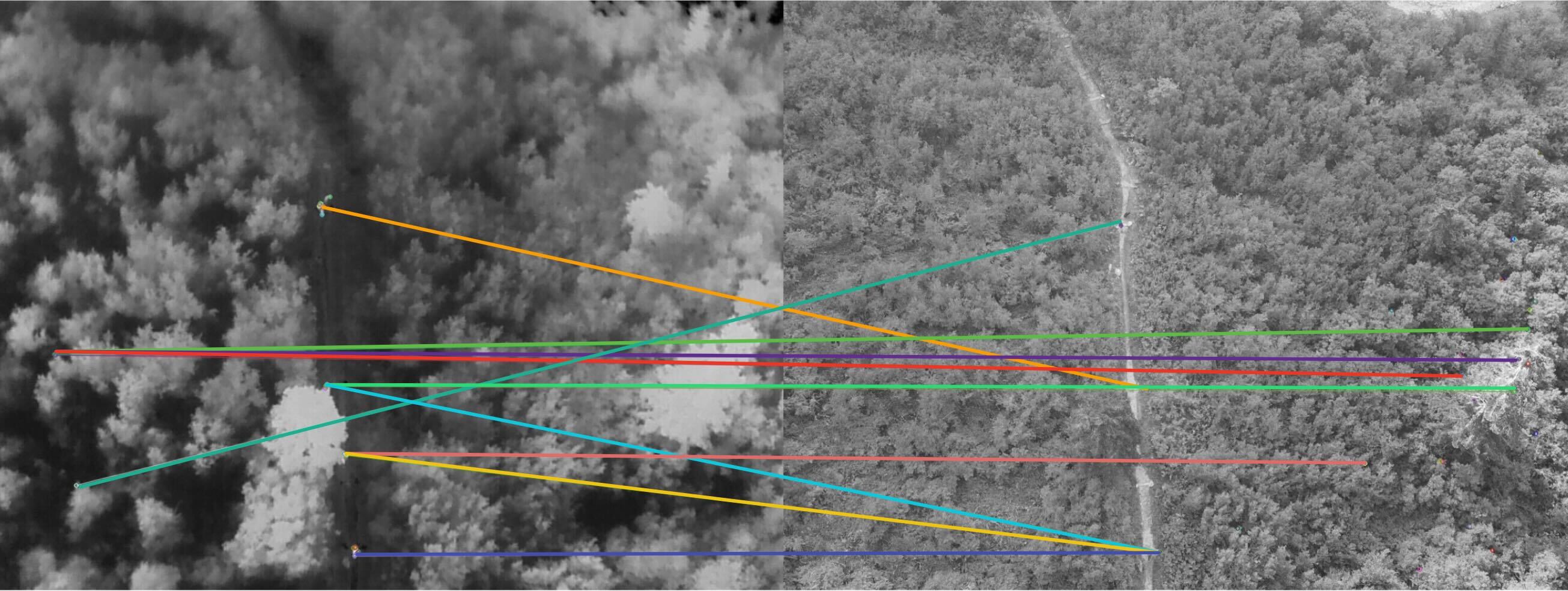}
   \caption{ORB-detected feature matches between thermal (left) and visual (right) images of a wilderness scene. The matches are mostly erroneous and unusable for image alignment.}
   \label{fig:registration difficulty blurry}
\end{figure}

\subsection{Visual-thermal image registration}
\label{subsec:registration challenge}
Multi-modal image registration on the WiSARD is particularly challenging because of the vastly different patterns across each modality that arise in different wilderness settings (see Figure~\ref{fig:image diversity} for examples).
Multi-modal image registration is the process of matching two or more images taken by different sensors and/or taken under different conditions (e.g., lighting, perspective) \cite{Li2017image}. Aligned thermal and visual images can be used in sensor fusion-based algorithms or used to assist a human with imagery analysis and reduce cognitive load \cite{Rasmussen2009-wq}, however in either case the alignment depends on the ability to identify common feature sets. Generating high-quality aligned or fused visual-thermal imagery is currently limited by the effectiveness of multi-spectral feature detection techniques. Classical image feature detection methods, such as the Scale Invariant Feature Transform (SIFT), Speed Up Robust Feature (SURF), and Oriented FAST and Rotated BRIEF (ORB), have been shown to be robust against a variety of transformations and distortions for visual imagery \cite{Karami2015image}, and ORB has been shown to be particularly good at detecting features in the long-wave infrared spectrum \cite{Ricaurte2014-uz}. However, for thermal-visual image pairs, these methods do not provide adequate feature matches required for registration. 
Figure \ref{fig:registration difficulty blurry} provides an example of ORB-detected features for a visual-thermal image pair in a wilderness setting. In this example, the trees appear blurry on the thermal image which makes feature matching with the visual image extremely challenging. Evidently, the feature matches are mostly erroneous and thus unusable for registration. 

Recent work leverages deep neural networks to address the cross-spectral alignment problem \cite{AchermannKolobovEtAl2020}. This method significantly outperforms classical methods for estimating the homography between multi-spectral image pairs in baseline tests, however the imagery used for training the neural network is qualitatively more benign (i.e., mostly farmland) compared to WiSARD. Additionally, the authors note there are still many cases in which all methods fail to correctly detect features and descriptors necessary for accurate alignment. We believe WiSARD provides many challenging scenes which would be useful for improving the accuracy and robustness of state-of-the-art approaches.

We understand that the registration problem for visual-thermal images is still an active area of research, and this example is to illustrate that the challenges are particularly pronounced in the WiSARD.
Thus, not only is the WiSARD a useful benchmark for object detection algorithms, but also for visual-thermal registration algorithms.

\section{Conclusion and Future Directions}
\label{sec:conclusion and future direction}
We are excited to release WiSARD, a set of annotated visual and thermal images collected from UAVs flying over human subjects in various wilderness terrains under diverse lighting, seasonal, and weather conditions. WiSARD offers a unique benchmark for testing human detection and sensor fusion algorithms in extremely challenging environments motivated by real-world WiSAR applications. To the best of our knowledge, WiSARD is the first large-scale multi-modal image dataset directed at the development of vision-based algorithms for autonomous WiSAR UAVs. Due to the diversity and various visual and thermal patterns found in wilderness contexts, we believe that this dataset will serve as a useful benchmark for testing the robustness of computer vision algorithms in real-world settings.

We plan on continuing building upon this dataset through a number of directions. First, we will gradually expand the size of the dataset and continue to work with first responders to ensure that the dataset remains practically relevant and diverse. In particular, we hope to expand the scope of the dataset to include disaster zones (e.g., floods, fires, earthquakes) as we envision that UAVs will be a useful asset in those settings too. Second, inspired by the set-up in \cite{ChoiKimEtAl2018}, we hope to improve upon our hardware configuration to make the raw visual and thermal images more aligned through hardware rather than software. Third, to provide momentum for research in this direction, we plan to develop and release a number of baseline algorithms for visual-thermal image registration and human detection.

\bibliographystyle{IEEEtran}
\bibliography{references}  

\end{document}

%% file: arxiv_cover_page.tex
%
%
%
%
%
%
%
\def \myJournal {IEEE International Conference on Intelligent Robots and Systems (IROS)}
\def \myDoi {}
\def \myPaperSiteName {IEEE Xplore}
\def \myPaperSiteLink {}
\def \myYear {2023}
\def \myPaperCitation{D. Broyles, C. R. Hayner and K. Leung, "WiSARD: A Labeled Visual and Thermal Image Dataset for Wilderness Search and Rescue," 2022 IEEE/RSJ International Conference on Intelligent Robots and Systems (IROS), Kyoto, Japan, 2022, pp. 9467-9474, doi: 10.1109/IROS47612.2022.9981298.}


\begin{figure*}[t]

\thispagestyle{empty}
\begin{center}
\begin{minipage}{6in}
\centering
This paper has been accepted for publication in \emph{\myJournal}. 
\vspace{1em}
This is the author's version of an article that has been published in this conference. Changes were made to this version by the publisher prior to publication.
\vspace{2em}


Please cite this paper as:

\myPaperCitation

\vspace{15cm}
\copyright 2022 \hspace{4pt}IEEE. Personal use of this material is permitted. Permission from IEEE must be obtained for all other uses, in any current or future media, including reprinting/republishing this material for advertising or promotional purposes, creating new collective works, for resale or redistribution to servers or lists, or reuse of any copyrighted component of this work in other works.

\end{minipage}
\end{center}
\end{figure*}
\newpage
\clearpage
\pagenumbering{arabic}

%% file: Broyles.Hayner.ea.IROS2022a.bbl
\begin{thebibliography}{10}
\providecommand{\url}[1]{#1}
\csname url@samestyle\endcsname
\providecommand{\newblock}{\relax}
\providecommand{\bibinfo}[2]{#2}
\providecommand{\BIBentrySTDinterwordspacing}{\spaceskip=0pt\relax}
\providecommand{\BIBentryALTinterwordstretchfactor}{4}
\providecommand{\BIBentryALTinterwordspacing}{\spaceskip=\fontdimen2\font plus
\BIBentryALTinterwordstretchfactor\fontdimen3\font minus \fontdimen4\font\relax}
\providecommand{\BIBforeignlanguage}[2]{{%
\expandafter\ifx\csname l@#1\endcsname\relax
\typeout{** WARNING: IEEEtran.bst: No hyphenation pattern has been}%
\typeout{** loaded for the language `#1'. Using the pattern for}%
\typeout{** the default language instead.}%
\else
\language=\csname l@#1\endcsname
\fi
#2}}
\providecommand{\BIBdecl}{\relax}
\BIBdecl

\bibitem{GeigerLenzEtAl2013}
A.~Geiger, P.~Lenz, C.~Stiller, and R.~Urtasun, ``{Vision meets robotics: The KITTI dataset},'' \emph{{Int.\ Journal of Robotics Research}}, vol.~32, no.~11, pp. 1231--1237, 2013.

\bibitem{waymo2019}
``{Waymo Open Dataset: An autonomous driving dataset},'' 2019, \url{https://www.waymo.com/open}.

\bibitem{nuscenes2019}
H.~Caesar, V.~Bankiti, A.~H. Lang, S.~Vora, V.~E. Liong, Q.~Xu, A.~Krishnan, Y.~Pan, G.~Baldan, and O.~Beijbom. (2019) {nuScenes: A multimodal dataset for autonomous driving}. {Available at }\url{https://arxiv.org/abs/1903.11027}.

\bibitem{TreibleSaponaroEtAl2017}
W.~Treible, P.~Saponaro, S.~Sorensen, A.~Kolagunda, M.~O’Neal, B.~Phelan, K.~Sherbondy, and C.~Kambhamettu, ``{CATS: A Color and Thermal Stereo Benchmark},'' in \emph{{IEEE Conf.\ on Computer Vision and Pattern Recognition}}, 2017.

\bibitem{ChoiKimEtAl2018}
Y.~Choi, N.~Kim, S.~Hwang, K.~Park, J.~S. Yoon, K.~An, and I.~S. Kweon, ``{KAIST Multi-Spectral Day/Night Data Set for Autonomous and Assisted Driving},'' \emph{{IEEE Transactions on Intelligent Transportation Systems}}, vol.~19, no.~3, pp. 934--948, 2018.

\bibitem{ZhanSunEtAl2019}
L.~Zhan, W.and~Sun, D.~Wang, H.~Shi, A.~Clausse, M.~Naumann, J.~Kummerle, H.~Konigshof, C.~Stiller, A.~Fortelle, and T.~M. (2019) {INTERACTION Dataset: An INTERnational, Adversarial and Cooperative moTION Dataset in Interactive Driving Scenarios with Semantic Maps}. {Available at }\url{https://arxiv.org/abs/1910.03088}.

\bibitem{Hwang2015peddet}
S.~Hwang, J.~Park, N.~Kim, Y.~Choi, and I.~S. Kweon, ``Multispectral pedestrian detection: Benchmark dataset and baseline,'' in \emph{{IEEE Conf.\ on Computer Vision and Pattern Recognition}}, 2015.

\bibitem{ZhuWenEtAl2021}
P.~Zhu, L.~Wen, D.~Du, X.~Bian, H.~Fan, Q.~Hu, and H.~Ling, ``{Detection and Tracking Meet Drones Challenge},'' \emph{{IEEE Transactions on Pattern Analysis \& Machine Intelligence}}, 2021.

\bibitem{GebhardtWolf2018}
E.~Gebhardt and M.~Wolf, ``{CAMEL Dataset for Visual and Thermal Infrared Multiple Object Detection and Tracking},'' in \emph{{IEEE Int.\ Conf.\ on Advanced Video and Signal Based Surveillance}}, 2018.

\bibitem{AchermannKolobovEtAl2020}
F.~Achermann, A.~Kolobov, D.~Dey, T.~Hinzmann, J.~Chung, R.~Siegwart, and N.~Lawrance, ``{MultiPoint: Cross-spectral registration of thermal and optical aerial imagery},'' in \emph{{Conf.\ on Robot Learning}}, 2020.

\bibitem{Bozic-StulicMarusicEtAl2019}
D.~Bo{\v{z}}i{\'{c}}-{\v{S}}tuli{\'{c}}, {\v{Z}}.~Maru{\v{s}}i{\'{c}}, and S.~Gotovac, ``{Deep Learning Approach in Aerial Imagery for Supporting Land Search and Rescue Missions},'' \emph{{Int.\ Journal of Computer Vision}}, vol. 127, no.~9, pp. 1256--1278, 2019.

\bibitem{MoralesVazquezEtAl2021}
J.~Morales, R.~V\'{a}zquez-Mart\'{i}n, A.~Mandow, D.~Morilla-Cabello, and A.~Garc\'{i}a-Cerezo, ``{The {UMA-SAR} Dataset: Multimodal Data Collection from a Ground Vehicle During Outdoor Disaster Response Training Exercises},'' \emph{{Int.\ Journal of Robotics Research}}, vol.~40, no. 6-7, pp. 835--847, 2021.

\bibitem{LinMaireEtAl2015}
T.~Lin, M.~Maire, S.~Belongie, L.~Bourdev, R.~Girshick, J.~Hays, R.~Perona, D.~Ramanan, C.~L. Zitnick, and P.~Doll\'{a}r, ``{Microsoft COCO: Common Objects in Context},'' in \emph{{European Conf.\ on Computer Vision}}, 2015.

\bibitem{DengDongEtAl2009}
J.~Deng, W.~Dong, R.~Socher, L.-J. Li, K.~Li, and L.~Fei-Fei, ``{ImageNet: A Large-Scale Hierarchical Image Database},'' in \emph{{IEEE Conf.\ on Computer Vision and Pattern Recognition}}, 2009.

\bibitem{OpenDataset2020}
A.~Kuznetsova, H.~Rom, N.~Alldrin, J.~Uijlings, I.~Krasin, J.~Pont-Tuset, S.~Kamali, S.~Popov, M.~Malloci, A.~Kolesnikov, T.~Duerig, and V.~Ferrari, ``{The Open Images Dataset V4: Unified image classification, object detection, and visual relationship detection at scale},'' \emph{{Int.\ Journal of Computer Vision}}, vol. 128, no.~7, pp. 1956--1981, 2020.

\bibitem{GeigerLenzEtAl2012}
A.~Geiger, P.~Lenz, and R.~Urtasun, ``{Are we ready for Autonomous Driving? The KITTI Vision Benchmark Suite},'' in \emph{{IEEE Conf.\ on Computer Vision and Pattern Recognition}}, 2012.

\bibitem{GoogleEarthImage}
G.~E.~P. v7.3.4.8248, ``Washington state,'' Accessed Feb. 28, 2022. Imagery from 1 Dec. 2020. 48$^\circ$00'49'' N, 121$^\circ$47'11.91'' W, elev 1481 ft, eye alt 198.28 mi, Image Landsat / Copernicus.

\bibitem{seibert1998flat}
J.~A. Seibert, J.~M. Boone, and K.~K. Lindfors, ``Flat-field correction technique for digital detectors,'' in \emph{Medical Imaging 1998: Physics of Medical Imaging}, vol. 3336.\hskip 1em plus 0.5em minus 0.4em\relax SPIE, 1998, pp. 348--354.

\bibitem{Gurton_undated-sg}
K.~P. Gurton and R.~Edmondson, ``{MidIR} and {LWIR} thermal polarimetric imaging comparison using receiver operating characteristic ({ROC}) curve analysis,'' \emph{{DEVCOM: Army Research Laboratory}}, 2020.

\bibitem{Zhao2016-gy}
H.~Zhao, Z.~Ji, N.~Li, J.~Gu, and Y.~Li, ``{Target Detection over the Diurnal Cycle Using a Multispectral Infrared Sensor},'' \emph{{Sensors}}, vol.~17, no.~1, 2016.

\bibitem{yolov5}
G.~Jocher and contributors. {YOLOv5}. {Available at }\url{https://github.com/ultralytics/yolov5}.

\bibitem{yolov3}
J.~Redmon and A.~Farhadi. (2018) {YOLOv3: An Incremental Improvement}. {Available at }\url{https://arxiv.org/abs/1804.02767}.

\bibitem{HaynerGuptaEtAl2021}
C.~R. Hayner, T.~Zhou, N.~Gupta, E.~Liu, P.~Mayhew, and J.~Vagners, ``{Real-time Human Detection with Integration of Visual and Thermal Data from High Altitude sUAS},'' in \emph{{AIAA Scitech Forum}}, 2021.

\bibitem{OzgeUnelBurakEtAl2019}
F.~\"{O}zge \"{U}nel, B.~O. \"{O}zkalayci, and C.~\c{C}i\v{g}la, ``{The Power of Tiling for Small Object Detection},'' in \emph{{IEEE Conf.\ on Computer Vision and Pattern Recognition}}, 2019.

\bibitem{Li2017image}
{Li, H. and Ding, W. and Cao, X. and Liu, C.}, ``{Image Registration and Fusion of Visible and Infrared Integrated Camera for {Medium-Altitude} Unmanned Aerial Vehicle Remote Sensing},'' \emph{{Remote Sensing}}, vol.~9, no.~5, 2017.

\bibitem{Rasmussen2009-wq}
N.~D. Rasmussen, B.~S. Morse, M.~A. Goodrich, and D.~Eggett, ``{Fused visible and infrared video for use in wilderness search and rescue},'' \emph{{IEEE Winter Conf.\ on Applications of Computer Vision}}, 2009.

\bibitem{Karami2015image}
E.~Karami, S.~Prasad, and M.~Shehata, ``{Image Matching Using {SIFT}, {SURF}, {BRIEF} and {ORB}: Performance Comparison for Distorted Images},'' in \emph{{Newfoundland Electrical and Computer Engineering Conference}}, 2015.

\bibitem{Ricaurte2014-uz}
P.~Ricaurte, C.~Chil{\'a}n, B.~X. Aguilera-Carrasco, C. A.and~Vintimilla, and A.~D. Sappa, ``{Feature point descriptors: infrared and visible spectra},'' \emph{{Sensors}}, vol.~14, no.~2, pp. 3690--3701, 2014.

\end{thebibliography}
